%% file: neurips_2026.tex
\documentclass{article}

\PassOptionsToPackage{numbers,sort&compress}{natbib}
\usepackage[preprint]{neurips_2026}
\usepackage[utf8]{inputenc}
\usepackage[T1]{fontenc}
\usepackage{hyperref}
\hypersetup{
    colorlinks=true,
}
\usepackage{url}
\usepackage{booktabs}
\usepackage{amsfonts}
\usepackage{amsmath}
\usepackage{amssymb}
\usepackage{nicefrac}
\usepackage{microtype}
\usepackage{xcolor}
\usepackage{graphicx}
\usepackage{adjustbox}
\usepackage{caption}
\usepackage{subcaption}
\usepackage{wrapfig}
\graphicspath{{figure/}}

\usepackage{booktabs}
\usepackage{multirow}
\usepackage{tabularx}
\usepackage{colortbl}
\usepackage{xcolor}
\usepackage{enumitem}
\usepackage[most]{tcolorbox}
 \usepackage{xcolor}             
\definecolor{ForestGreen}{RGB}{34,139,34}              
\definecolor{BrickRed}{RGB}{178,34,34}

\newcommand{\promptvar}[1]{\textcolor{blue!55!black}{\textit{#1}}}
\newcommand{\promptsec}[1]{\textbf{\color{black!75}#1}}

\newtcolorbox{promptbox}[1]{%
    enhanced,
    colback=white, colframe=gray!60!black,
    boxrule=0.4pt, arc=2pt,
    left=10pt, right=10pt, top=8pt, bottom=8pt,
    fonttitle=\bfseries\small\color{white},
    coltitle=white,
    colbacktitle=gray!55!black,
    title=#1,
    attach boxed title to top left={yshift=-2mm,xshift=4mm},
    boxed title style={colback=gray!55!black, sharp corners,
                       boxrule=0pt, top=2pt, bottom=2pt,
                       left=6pt, right=6pt}}
                       
\title{Continuous Audio Thinking for\\ Large Audio Language Models}

\author{
    \textbf{Gyojin Han$^{1}$\thanks{Equal contribution.} \quad Dong-Jae Lee$^{1*}$ \quad Changho Choi$^{1*}$ \quad Jongsuk Kim$^{2*}$ \quad Junmo Kim$^{1}$} \\
    $^{1}$KAIST, South Korea \quad\quad $^{2}$PYLER, South Korea \\
    {\texttt{\small \{hangj0820, jhtwosun, ccho4702, junmo.kim\}@kaist.ac.kr}, \quad\quad \texttt{\small js.kim@pyler.tech}} \\
\vspace{-20pt}
}

\begin{document}

\maketitle

\begin{abstract}
Large audio language models (LALMs) have shown impressive capabilities on diverse audio understanding tasks, ranging from speech transcription to music analysis. However, because LALMs are typically trained to produce text-aligned responses, their hidden states are progressively shaped for text generation rather than for preserving acoustic information. As a result, the diverse acoustic content that audio carries, such as phonetic detail, prosody, sound events, affect, and pitch, is lost along the way and difficult to leverage in the response. We introduce Continuous Audio Thinking (CoAT), a framework that equips audio language models with a continuous latent workspace for organizing acoustic information prior to response generation, grounded by distillation from audio experts. Within the thinking space, the model can utilize the rich acoustic information provided by expert distillation when generating its response. Furthermore, the proposed continuous thinking block can be processed in a single prefill, so CoAT does not require additional autoregressive decoding cost over the baseline. Across three LALMs, Qwen2-Audio, Qwen2.5-Omni-7B, and Audio Flamingo~3, performance gains on a broad benchmark suite spanning audio reasoning, audio understanding, music classification, speech emotion, and speech transcription demonstrate the effectiveness of CoAT. Further analysis confirms that the auxiliary supervision propagates from the thinking positions to the model's textual responses.
\end{abstract}

\input{tex/1_intro.tex}
\input{tex/2_rel.tex}
\input{tex/3_method.tex}

\input{tex/4_exp.tex}
\input{tex/5_con.tex}

\bibliography{neurips_2026}

\ \
\newpage
\input{tex/appendix}

\end{document}

%% file: tex/1_intro.tex
\section{Introduction}
\label{sec:introduction}

Large Audio Language Models (LALMs)~\citep{qwen2audio, qwen25omni,audioflamingo3, salmonn} have established themselves as a natural interface for understanding speech, environmental sounds, music, and other acoustic signals through language.
These models couple an audio encoder with a language model trained to autoregressively generate textual responses.
This design enables strong progress on spoken dialogue and audio question answering, but it also introduces a fundamental supervision mismatch.
The input contains rich frame-level acoustic structure, while the training objective provides signal only through sparse response tokens. 
The layers above the audio encoder are encouraged to retain only the information that is immediately useful for predicting the next text token, leaving many fine-grained acoustic cues weakly supervised or discarded.

This mismatch is especially pronounced for audio, which carries many properties that transcription alone cannot convey, including phonetic detail, speaker affect, background scene, prosody, and musical structure. One possible solution is to have the model verbalize intermediate thinking, as in discrete text chain-of-thought prompting~\citep{wei2022chain, kojima2022large, wang2022self}, describing what the audio contains in natural language (for instance, ``a man speaking in a high-pitched voice'' or ``the first note is a C'') before producing the answer. 
However, many acoustic attributes cannot be serialized into natural language without losing fine-grained temporal and spectral detail. Faithful natural-language rationales for them are rarely available at scale, and even when they are, compressing low-level acoustic evidence into text introduces an unnecessary bottleneck.

\begin{figure}
    \centering
    \includegraphics[width=\linewidth]{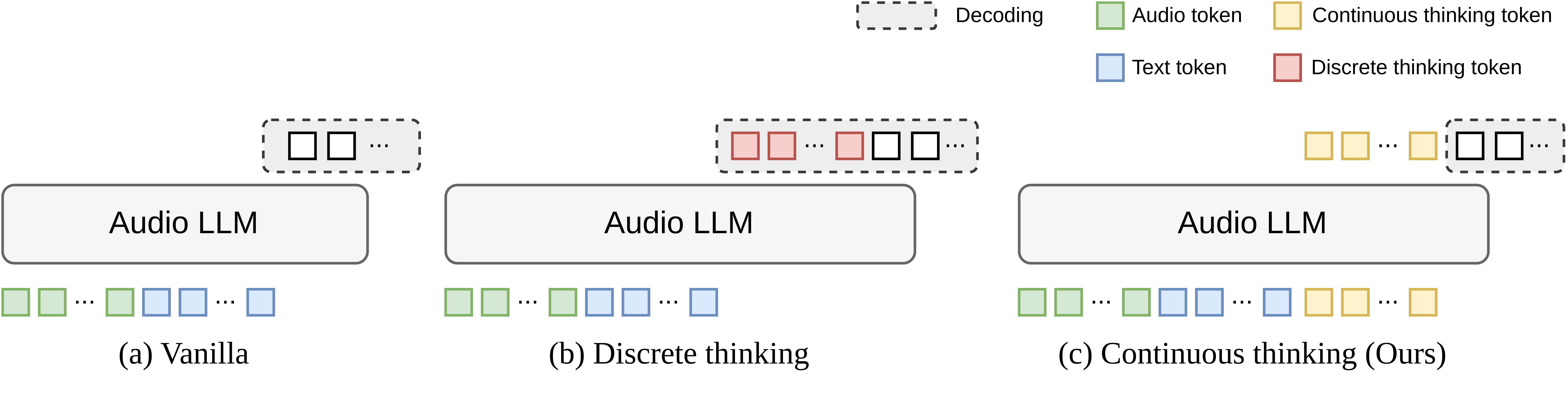}
    \caption{\textbf{Thinking paradigms in audio language models.}
    (a) Vanilla audio LMs decode the response directly from audio and instruction tokens.
    (b) Discrete thinking generates textual thinking tokens autoregressively before the answer.
    (c) Continuous Audio Thinking (ours) prepends a fixed-length block of continuous thinking tokens that is consumed in a single prefill, letting the model think in an audio-aligned latent space without autoregressive cost.}
    \label{fig:intro}
\end{figure}
These limitations motivate a new form of thinking that preserves and reorganizes acoustic information itself rather than verbalizing it. 
To meet this need, such thinking is required to unfold in continuous latent positions, remaining independent of text and free from supervision by natural-language rationales. It is further expected to serve as a workspace in which the language model maintains, aligns, and transforms acoustic information before committing to response generation.

Toward this, we propose \textbf{Continuous Audio Thinking} (\textbf{CoAT}), an auxiliary-supervision framework that equips audio language models with a latent workspace.
CoAT places a thinking block between the user input and the assistant response, where the model organizes acoustic information before generating its reply. 
The thinking block is grounded by distillation from diverse audio experts covering reconstruction, speech content~\citep{poli2025spidr}, sound events~\citep{kong2020panns}, paralinguistic features~\citep{ma2024emotion2vec}, and pitch~\citep{basicpitch}, providing complementary acoustic dimensions that text supervision alone cannot supply.
CoAT integrates with existing LALMs without architectural changes. Furthermore, CoAT requires neither textual rationales nor task-specific decoding formats, and can be learned from a modest amount of audio data.  
At inference, the thinking block is consumed in a single prefill, adding no autoregressive decoding cost over the baseline as shown in Figure~\ref{fig:intro}.

We instantiate CoAT on three LALMs (Qwen2-Audio~\citep{qwen2audio}, Qwen2.5-Omni-7B~\citep{qwen25omni}, and Audio Flamingo 3~\citep{audioflamingo3}) and evaluate it across diverse audio understanding and reasoning benchmarks. 
CoAT yields consistent gains across backbones and outperforms text chain-of-thought with substantially lower per-sample latency. 
Further analysis confirms that the auxiliary supervision propagates from the thinking positions to the model's textual responses.
We summarize our contributions as follows:

\begin{itemize}
    \item We propose a continuous-thinking paradigm for LALMs, enabling the model to organize acoustic information in the latent space without verbalization or autoregressive decoding.
    \item We propose a multi-expert distillation objective that grounds the thinking states in complementary acoustic dimensions, supplying signal that text supervision alone cannot provide.
    \item We evaluate CoAT on three LALMs and show that it consistently improves audio understanding and reasoning at lower latency than text chain-of-thought, with analysis showing that the supervision propagates from the thinking positions to the model's textual responses.
\end{itemize}

%% file: tex/2_rel.tex
\section{Related Work}
\label{sec:rel}

\paragraph{Audio Language Models.}
A growing family of large language models has been adapted to natively process audio alongside text~\citep{qwenaudio, qwen2audio, qwen25omni, audioflamingo, audioflamingo2, audioflamingo3, salmonn, ltu, pengi, gama, llasm, mullama, anygpt, onellm}.
A common architecture connects a pretrained audio encoder, typically an audio-text-aligned Whisper~\citep{whisper} encoder, to a language model via a lightweight projection layer, exposing audio as a stream of soft-token embeddings that are consumed by the decoder~\citep{salmonn, audioflamingo, audioflamingo2, audioflamingo3, ltu, pengi, gama}.
Another line trains a single multimodal model end-to-end so that audio, vision, and text share a unified token space~\citep{qwen25omni, anygpt, onellm}.
In both cases, audio capabilities are typically obtained from supervised fine-tuning on triplets of audio, instruction, and target response, sometimes structured as a multi-stage curriculum that first aligns the audio encoder before instruction-tuning the language model~\citep{audioflamingo3, qwen25omni}.

\paragraph{Continuous and Latent Thinking.}
Chain-of-thought prompting~\citep{wei2022chain, kojima2022large, wang2022self} demonstrated that explicit step-by-step traces in language tokens improve the reasoning behavior of large language models.
Subsequent work has examined whether reasoning must remain in the discrete token space: Coconut~\citep{hao2024training} replaces textual chains with continuous embeddings that are looped back into the model, and Quiet-STaR~\citep{zelikman2024quiet} learns implicit thoughts that are scored against the next-token likelihood.
A related thread treats internal recurrence and depth as an explicit reasoning resource via looped transformers and implicit chain-of-thought formulations~\citep{giannou2023looped, deng2024explicit}.
A concurrent line extends continuous reasoning beyond text: Chain-of-Visual-Thought~\citep{qin2025chain} introduces continuous visual tokens that allow vision-language models to reason in a vision-aligned latent space, but it still autoregressively generates additional tokens for each task and pairs them with explicit textual reasoning, leaving the inference-cost limitations of token-by-token reasoning largely unaddressed.

\paragraph{Audio Encoders for Diverse Tasks.}
A wide range of audio encoders has been developed, each capturing a different aspect of the signal.
Speech understanding is supported by self-supervised speech models that learn linguistic units from raw waveforms~\citep{hsu2021hubert, baevski2020wav2vec, chen2022wavlm, poli2025spidr}.
General sound-event analysis builds on classifiers and masked models pretrained on broad acoustic taxonomies~\citep{kong2020panns, chen2023beats, huang2022masked, chen2024eat}.
Music understanding is addressed by encoders specialized for tonal and rhythmic structure~\citep{yizhimert, basicpitch}, while paralinguistic analysis relies on self-supervised models trained on emotional speech~\citep{ma2024emotion2vec}.
Neural audio codecs provide a complementary representation, compressing waveforms into compact latent sequences from which the original signal can be decoded~\citep{zeghidour2021soundstream, defossez2022high, simwhisper}.
We collect these complementary encoders into a single place and jointly utilize them in training time to enable an audio language model to think continuously, turning a scattered set of task-specific representations into a single audio-aware reasoning substrate.

%% file: tex/3_method.tex
\section{Method}
\label{sec:method}
\begin{figure}[t]
    \centering
    \includegraphics[width=\linewidth]{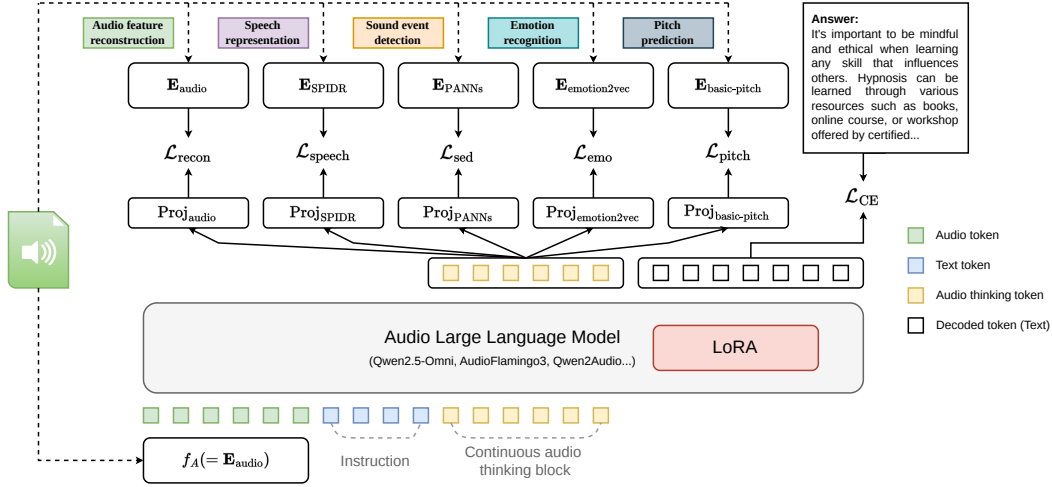}
    \caption{\textbf{CoAT architecture.}
    A continuous audio thinking block is supervised by five audio experts via per-task projection heads, covering audio feature reconstruction, speech representation, sound event detection, paralinguistic features, and pitch. The projection heads decode the shared hidden states into expert-aligned predictions, used only during training.}
    \label{fig:method}
\end{figure}

In this section, we propose Continuous Audio Thinking (CoAT), a method that allows LALMs to retain and organize acoustic content in latent form before generating their textual response. 
CoAT inserts a sequence of continuous latent positions between the audio input and the assistant response, and supervises the hidden states at those positions by distillation against multiple audio experts. 
The thinking block thus serves as a workspace grounded in complementary acoustic dimensions that text supervision alone cannot provide. 
We first formalize the audio language model interface (\S\ref{sec:audio_lm}), then introduce the thinking block (\S\ref{sec:architecture}), the expert distillation (\S\ref{sec:tasks}), and the stage-wise training schedule (\S\ref{sec:objective}).
The overall pipeline of the proposed method is shown in Figure~\ref{fig:method}.

\subsection{Large Audio Language Models}
\label{sec:audio_lm}

Before describing the method, we briefly outline how LALMs operate.
We consider an audio language model composed of an LLM decoder $f_L$ and an audio encoder $f_A$. 
The encoder $f_A$ encodes a raw audio waveform $a$ into a sequence of $L_a$ audio token embeddings $\mathbf{x}_A \in \mathbb{R}^{L_a \times d}$ at a fixed rate $r_s$, where $d$ is the hidden size of $f_L$.
The input sequence is composed of a system prompt and a user prompt, $\mathbf{x} = [\mathbf{x}^{\text{sys}}, \mathbf{x}^{\text{usr}}]$, where the user prompt itself contains the audio token embeddings followed by a text prompt, $\mathbf{x}^{\text{usr}} = [\mathbf{x}_A, \mathbf{x}_{\text{txt}}]$.
The model is trained with next-token prediction using the cross-entropy loss
\begin{equation}\label{eq:ce}
\mathcal{L}_{\text{CE}}(\mathbf{x}) = -\sum_{t \in \mathcal{I}(\mathbf{x})} \log p_{f_L}(x_t \mid x_{<t}),
\end{equation}
where $\mathcal{I}(\mathbf{x})$ indexes positions in the assistant response.

\subsection{Continuous Audio Thinking Block}
\label{sec:architecture}
The thinking block introduces dedicated capacity for the model to process acoustic information before generating its response. 
We extend the model vocabulary with three special tokens, $\tau_s = \texttt{<|audio\_think\_start|>}$, $\tau_p = \texttt{<|audio\_think|>}$, and $\tau_e = \texttt{<|audio\_think\_end|>}$, that serve as the boundary and content positions of an audio thinking block.
Given an input that contains an audio segment of length $L_a$, the corresponding thinking block is constructed by placing one $\tau_p$ for each audio token, framed by the boundary tokens,
\begin{equation}\label{eq:block}
\mathbf{b}(L_a) = \big[\, \tau_s, \, \underbrace{\tau_p, \ldots, \tau_p}_{L_a}, \, \tau_e \,\big].
\end{equation}
The block $\mathbf{b}(L_a)$ is appended to the input sequence, $\tilde{\mathbf{x}}=[\mathbf{x}^{\text{sys}}, \mathbf{x}^{\text{usr}}, \mathbf{b}(L_a)]$, and processed by decoder $f_L$. 
We collect the final-layer hidden states $\mathbf{H}_{\text{think}} \in \mathbb{R}^{L_a \times d}$ at the positions $\tau_p$.
At training time, the cross-entropy in Eq.~\eqref{eq:ce} is computed only at response positions; the thinking-block tokens ${\tau_s, \tau_p, \tau_e}$ are excluded as next-token prediction targets.
During inference, the same block is appended deterministically before generation, adding prefill cost without additional autoregressive decoding.

\subsection{Distillation from Audio Experts}
\label{sec:tasks}

The thinking block provides space for processing acoustic information, but the text-only objective alone is insufficient for the model to learn to use this space effectively. 
We therefore distill frame-level features from audio experts to fill this gap.
Let $K$ denote the set of frozen audio experts used in CoAT. For each expert $k \in K$, an encoder $\textbf{E}_k$ produces expert features $\mathbf{z}_k = \textbf{E}_k(a) \in \mathbb{R}^{L_k \times e_k}$ from the raw audio $a$.
For each expert $k \in K$, we attach a projection head $P_k$ to the language model that projects $\mathbf{H}_{\text{think}}$ into expert-aligned predictions,
\begin{equation}\label{eq:proj}
\hat{\mathbf{z}}_k = P_k(\mathbf{H}_{\text{think}}) \in \mathbb{R}^{L_a \times e_k}.
\end{equation}
Each $P_k$ is implemented as a single-block Transformer (multi-head attention and a feed-forward layer) followed by a linear map to the expert embedding dimension $e_k$. The experts fall into two families: representational experts and task-specialized experts, described in turn below.

\subsubsection{Representational experts for location and content}
\label{sec:tasks:representational}

Representational experts, comprising audio feature reconstruction and speech distillation, establish a foundation on which all subsequent supervision rests.
Audio feature reconstruction anchors $\mathbf{H}_{\text{think}}$ to the audio encoder's latent space, so that the thinking block occupies the same subspace as the input acoustic representation.
Speech distillation then enriches that representation with the linguistic structure captured by a self-supervised speech encoder, endowing the thinking block with both location and content before any application-specific objective is introduced.

\paragraph{Audio Feature Reconstruction.}
The expert encoder is the audio encoder of the backbone itself, $\mathbf{E}_{\text{audio}}=f_A$.
The thinking block is trained to reproduce the same latent representation that $f_L$ consumes for the audio, which constrains $\mathbf{H}_{\text{think}}$ to the audio-token subspace and supervises where acoustic information should be encoded before any semantic objective is introduced.
We supervise this with a frame-wise MSE, $\mathcal{L}_{\text{recon}} = \mathrm{MSE}(\hat{\mathbf{z}}_{\text{audio}}, \mathbf{z}_{\text{audio}})$.

\paragraph{Speech Representation Distillation.}
Furthermore, we employ $\mathbf{E}_\text{SPIDR}$~\citep{poli2025spidr}, a self-supervised speech encoder trained without labels to extract stable linguistic units from raw waveforms.
We use its encoder output as the expert feature, supervised with a frame-wise MSE, $\mathcal{L}_{\text{speech}} = \mathrm{MSE}(\hat{\mathbf{z}}_{\text{SPIDR}}, \mathbf{z}_{\text{SPIDR}})$.
This task aligns the thinking representation with phonetic and lexical content, complementing the surface-level signal from audio feature reconstruction.

\subsubsection{Task-specialized experts for application-domain capabilities}
\label{sec:tasks:specialized}

Building on this foundation, we attach experts that supply task-specialized capabilities.
Each adds a capability that is otherwise difficult to acquire under text supervision alone, namely sound event detection for environmental audio, paralinguistic features for vocal affect, and pitch prediction for harmonic and prosodic structure.
Importantly, this design is not specific to the three experts used here. Any audio encoder that captures a desired representation can be incorporated as an additional task.

\paragraph{Sound Event Detection.}
For sound-event semantics, we use the expert encoder $\mathbf{E}_{\text{PANNs}}$~\citep{kong2020panns}, a CNN-based audio tagger pre-trained on AudioSet that produces frame-wise activations over 527 sound-event classes spanning speech, music, animal, vehicle, and ambient categories.
We map the student prediction into the same class-logit space by passing it through PANNs's final classification head $g_{\mathrm{cls}}$, and match it to the expert's per-frame, per-class probabilities with binary cross-entropy, $\mathcal{L}_{\text{sed}} = \mathrm{BCE}(g_{\mathrm{cls}}(\hat{\mathbf{z}}_{\text{PANNs}}), g_{\mathrm{cls}}(\mathbf{z}_{\text{PANNs}}))$.
This task exposes the thinking representation to a broad taxonomy of sound-event semantics, supporting general audio understanding tasks beyond speech.

\paragraph{Paralinguistic Feature Prediction.}
Voice affect is captured by $\mathbf{E}_\text{emotion2vec}$~\citep{ma2024emotion2vec}, a self-supervised model trained on speech emotion data that produces frame-wise representations.
We use its hidden-state output as the expert feature, supervised with a frame-wise MSE, $\mathcal{L}_{\text{emo}} = \mathrm{MSE}(\hat{\mathbf{z}}_{\text{emotion2vec}}, \mathbf{z}_{\text{emotion2vec}})$.
This task adds a non-lexical channel to the thinking representation that captures how the audio is spoken, such as affect, prosody, and intensity, rather than what is said.

\paragraph{Pitch Prediction.}
For pitch and harmonic structure, we adopt $\mathbf{E}_\text{basic-pitch}$~\citep{basicpitch}, a polyphonic pitch detector pre-trained on instrument transcription with a multi-pitch posteriorgram output.
We use its intermediate convolutional activations as the expert feature; the dense intermediate representation preserves harmonic information while avoiding the sparsity of the final posteriorgram.
The pitch task combines two losses, an MSE term on the intermediate feature and an auxiliary focal-BCE on the posteriorgram obtained by passing both student and expert features through the encoder's final convolution $h$. The loss is given by $\mathcal{L}_{\text{pitch}} = \mathrm{MSE}(\hat{\mathbf{z}}_{\text{basic-pitch}}, \mathbf{z}_{\text{basic-pitch}}) + w_{\text{pitch}}^{\text{aux}}  \mathrm{focal\text{-}BCE}(h(\hat{\mathbf{z}}_{\text{basic-pitch}}), h(\mathbf{z}_{\text{basic-pitch}}))$.
Pure BCE on the highly sparse posteriorgram collapses to all-zero predictions, while pure MSE on the dense intermediate feature under-constrains the pitch contour. 
This task introduces fine-grained pitch information into the thinking representation, which is otherwise sparsely covered by the speech- and sound-centric experts.

\subsection{Training Objective}
\label{sec:objective}

\paragraph{Stage-Wise Training.}
Training is organized as a sequence of $S$ stages.
Each stage is defined by its data, step budget, learning rate, and an active task subset $\mathcal{A}_s \subseteq K$ with per-task weights $w_k^{(s)}$, where $s$ denotes the stage index.
Each stage applies only the expert distillation tasks in $\mathcal{A}_s$ in addition to the language-modeling cross-entropy.
The parameter scope spans LoRA~\cite{lora} adapters, projection heads, added tokens, and the embedding layer.
We instantiate $S = 2$ in our main experiments, with a warm-up stage $\mathcal{A}_1$ using audio feature reconstruction alone, followed by a multi-task stage with $\mathcal{A}_2$ spanning all experts.

\paragraph{Overall Objective.}
The total loss at stage $s$ is the sum of the language-modeling and distillation losses
\begin{equation}\label{eq:total}
\mathcal{L}^{(s)}_{\text{total}}(\tilde{\mathbf{x}}) = \mathcal{L}_{\text{CE}}(\tilde{\mathbf{x}}) + \sum_{k \in \mathcal{A}_s} w_k^{(s)} \, \mathcal{L}_k,
\end{equation}
with $\mathcal{L}_{\text{CE}}$ computed on $\tilde{\mathbf{x}}$ under the thinking-token mask.
The thinking block thus contributes purely through the distillation losses, leaving the model's text generation unchanged.

%% file: tex/4_exp.tex
\section{Experiments}
\label{sec:exp}

\subsection{Experimental Setup}
\label{sec:impl}

We instantiate CoAT on three pretrained audio language models: Qwen2-Audio~\citep{qwen2audio}, Qwen2.5-Omni-7B~\citep{qwen25omni}, and Audio Flamingo 3~\citep{audioflamingo3}. 
All backbones are trained with the same two-stage CoAT schedule: a reconstruction-only warm-up stage followed by a multi-task stage with all five experts active. 

We use the same public training mixture, sampling policy, and evaluation protocol across all backbones and benchmarks. The training mixture covers automatic speech recognition~\citep{librispeech, gigaspeech, commonvoice, voxpopuli, switchboard, spgispeech}, audio and speech question answering~\citep{audiocaps, iemocap, librisqa, clothoaqa}, audio captioning~\citep{audiocaps, clotho}, multiple-choice question audio understanding~\citep{meld, iemocap, librisqa}, music understanding~\citep{musicbench}, spoken-instruction following~\citep{gsqa}, and a small text-only supervised fine-tuning split~\citep{wildjailbreak}. Further experimental details are provided in Appendix~\ref{app:configs}.

\subsection{Main Results}
\label{sec:exp:main}

\input{table/main_all}

Across all three backbones, CoAT improves the pretrained baseline on the majority of benchmarks, as reported in Table~\ref{tab:main_all}, demonstrating its generality across various models and tasks.
The improvements are most pronounced on understanding- and reasoning-intensive tasks such as MELD, MMAR, and Alpaca-Audio, indicating that CoAT is particularly effective on audio understanding and reasoning
On automatic speech recognition, CoAT substantially improves the weaker Qwen2-Audio backbone while preserving performance on Qwen2.5-Omni and Audio Flamingo 3.
Overall, CoAT generalizes across heterogeneous audio-language backbones and yields the largest gains on reasoning-heavy audio understanding.

\subsection{Comparison with Discrete Thinking}
\label{sec:exp:discrete}

We compare CoAT's continuous thinking approach against the discrete-thinking alternative where the model autoregressively generates a textual reasoning trace before producing the answer. For Audio Flamingo 3, which has a native think model, we use built-in chain-of-thought template. For Qwen2.5-Omni, which has no native think mode, we instead induce text-CoT through a prompt-level instruction asking the model to reason step by step.

Table~\ref{tab:comp_discrete} reports reasoning accuracy on two benchmarks (MMAU and MMAR) with four inference-cost metrics: time to first token (TTFT), decoding time (Dec. time), the number of decoded tokens (Dec. tok), and end-to-end latency (Total).
Each reported value is the mean over multiple runs measured in the same environment; full experimental details are provided in Appendix~\ref{app:speed-duration}.

\input{table/comp_discrete}

Text-CoT and CoAT differ in where they spend inference cost. Text-CoT autoregressively decodes a reasoning chain before the answer, so its overhead lands almost entirely in the decode stage and does not amortize across batched requests. CoAT instead consumes its thinking block in a single prefill, shifting the overhead out of the decoding phase. As a result, CoAT runs much faster than text-CoT while improving reasoning accuracy over the same-backbone baseline. Appendix~\ref{app:af3-think} provides a full per-benchmark comparison with Audio Flamingo 3's native think mode.

\subsection{Analysis}
\label{sec:exp:analysis}
\begin{wrapfigure}{r}{0.5\textwidth}
  \centering
  \vspace{-1.4em}
  \includegraphics[width=\linewidth]{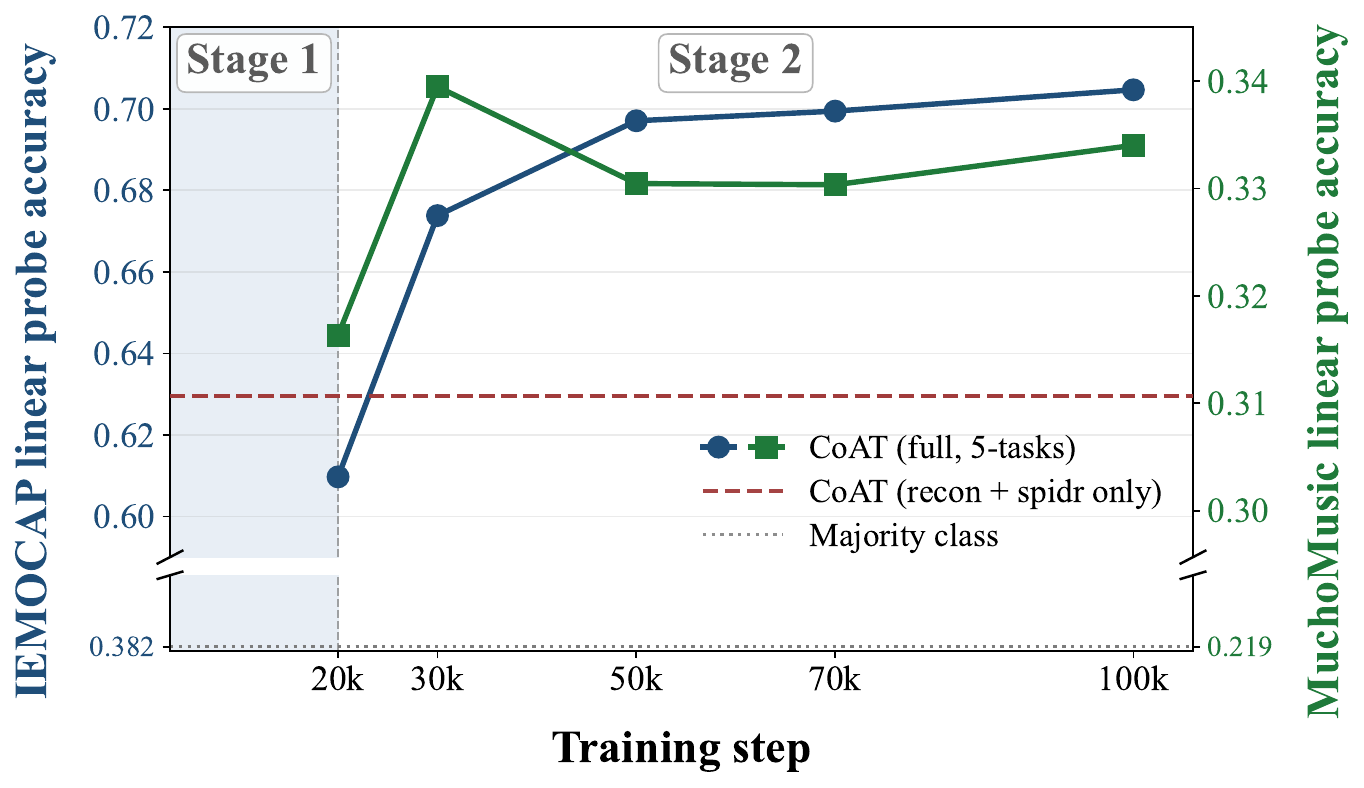}
  \vspace{-1.5em}
  \caption{Linear probe accuracy at the audio-think hidden across training checkpoints, on 4-class IEMOCAP emotion and 12-class MuchoMusic dominant pitch.}
  \vspace{-2.5em}
  \label{fig:probe_trajectory}
\end{wrapfigure}
We probe whether CoAT's auxiliary supervision injects task-relevant information at thinking positions by training linear probes on the LM hidden state at two positions.
The first one is the audio-think hidden, taken as the mean over the thinking block $\tau_p$ where CoAT's supervision attaches.
The second one is the pre-generation hidden, taken from the last $\tau_e$ token for CoAT and from the last input token before the assistant turn for the vanilla baseline. Both correspond to the model's decision-time representation.

\paragraph{Linear probing accuracy across training checkpoints.}
We probe two targets aligned with the training supervision, specifically 4-class emotion on IEMOCAP and 12-class dominant pitch on MuchoMusic.
Linear probes are trained on the audio-think hidden of the full 5-head CoAT across training checkpoints.
Figure~\ref{fig:probe_trajectory} shows that probe accuracy rises in stage 2 on both targets, when specialized experts begin contributing supervision beyond reconstruction.
This indicates that the auxiliary supervision injects task-relevant information into the supervised position.

\paragraph{Correlation between thinking-token information and downstream performance.}
\begin{wraptable}{r}{0.5\textwidth}
  \vspace{-1.3em}
  \caption{Probe accuracy and Spearman $\rho$ between probe confidence and downstream task performance, reported as ($\text{accuracy}/\rho$). \textit{Think} and \textit{Pre-gen} denote the audio-think and pre-generation hidden states, respectively.}
  \label{tab:probe_mediation}
  \input{table/probe_mediation}
  \vspace{-1em}
\end{wraptable}
We now probe two targets aligned with the downstream task, namely 4-class emotion on IEMOCAP and 7-class instrument family on the MuchoMusic instrument-question subset.
We compare three models: the vanilla Qwen2.5-Omni baseline, a CoAT control learned only by representational experts, and the CoAT trained with all five experts.
Table~\ref{tab:probe_mediation} reports probe accuracy alongside within-model Spearman correlations between probe confidence and downstream task performance.
The CoAT trained with all five experts attains the highest probe accuracy and the strongest within-model correlation across all cases.
Together, the supervision injects information at the thinking position that accumulates over training and correlates with downstream task performance. 

\begin{figure}
    \centering
    \captionsetup[sub]{skip=0pt}
    \begin{subfigure}[b]{0.49\linewidth}
        \centering
        \includegraphics[width=\linewidth]{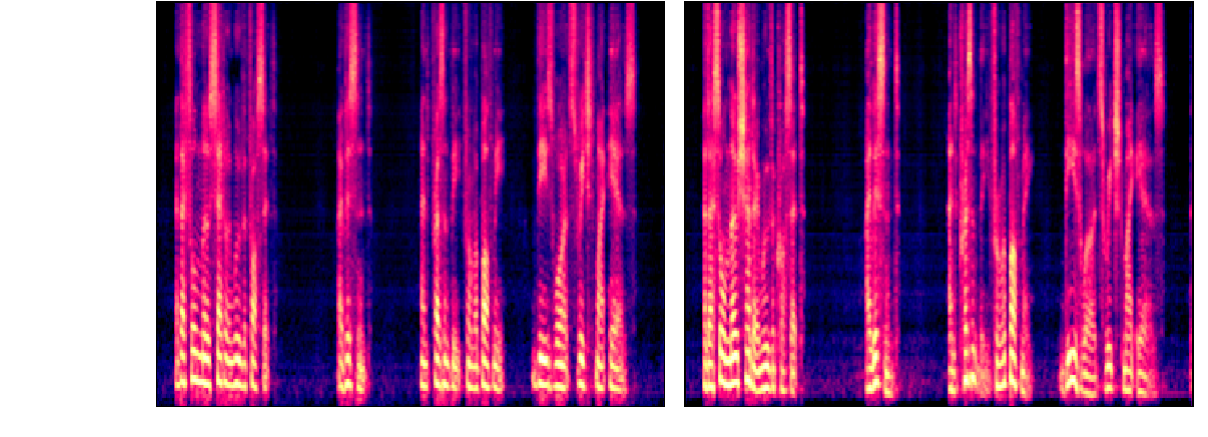}
        \caption{Audio feature reconstruction}
    \end{subfigure}
    \hfill
    \begin{subfigure}[b]{0.49\linewidth}
        \centering
        \includegraphics[width=\linewidth]{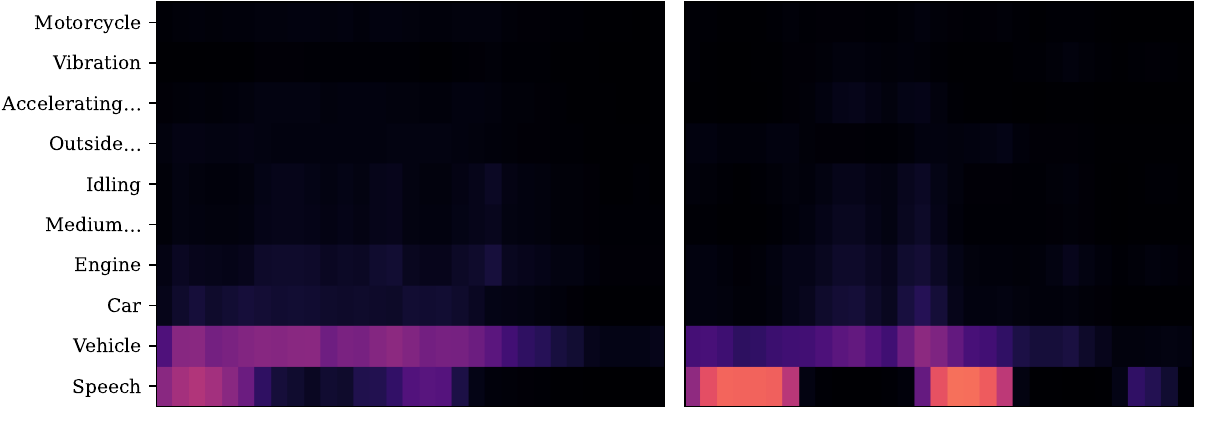}
        \caption{Sound event detection}
    \end{subfigure}
    \\[0.2em]
    \begin{subfigure}[b]{0.49\linewidth}
        \centering
        \includegraphics[width=\linewidth]{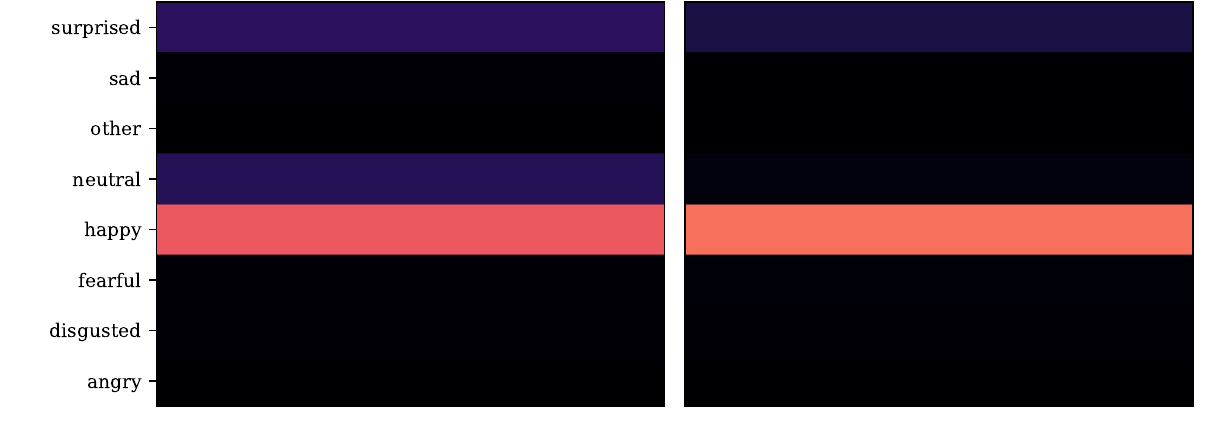}
        \caption{Paralinguistic feature prediction}
    \end{subfigure}
    \hfill
    \begin{subfigure}[b]{0.49\linewidth}
        \centering
        \includegraphics[width=\linewidth]{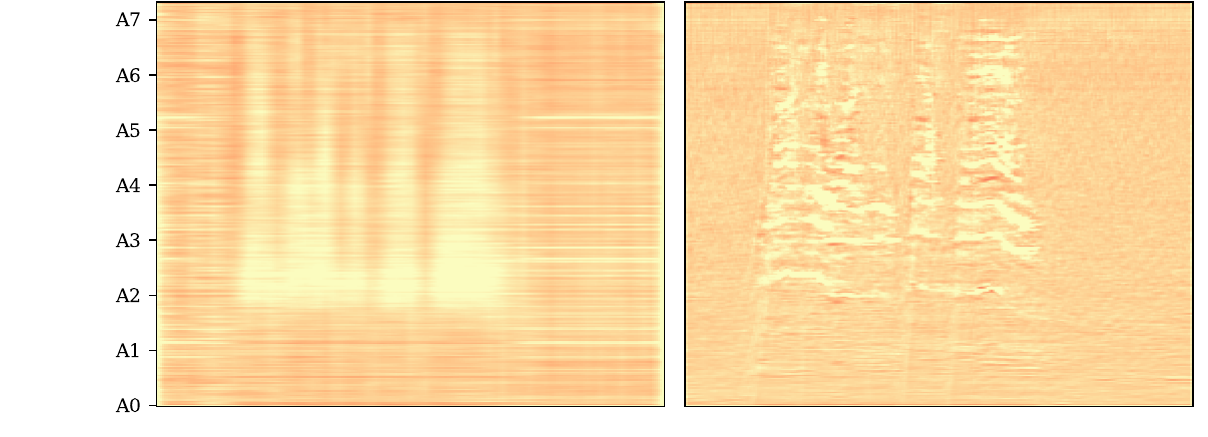}
        \caption{Pitch prediction}
    \end{subfigure}
    \caption{\textbf{Example reconstructions from CoAT's per-task heads.} Each pair shows the expert target (right) and the corresponding student prediction (left) at the audio-think positions.}
    \label{fig:vis}
\end{figure}

\paragraph{Visualizing per-head reconstructions.}
To verify that the thinking representation distills information from each expert, we visualize example reconstructions in Figure~\ref{fig:vis}.
For audio feature reconstruction, we use the Sim-Whisper~\citep{simwhisper} codec decoder to reconstruct audio from the predicted feature, which we visualize as a mel-spectrogram.
For the other tasks, we pass the predicted feature through the expert's prediction head and visualize the resulting output.
Each task's reconstruction faithfully matches the expert target, confirming that the thinking representation encodes the information required to perform every supervised task.

\subsection{Ablation Studies}
\label{sec:abl}

We conduct ablations on a single backbone, Qwen2.5-Omni, to validate our design choices.

\paragraph{Task Ablation.}

Table~\ref{tab:abl_task} reports the cumulative contribution of each component. 
Supervised fine-tuning on top of Qwen2.5-Omni substantially improves Emotion and reduces ASR error, but leaves General and AIR-Bench essentially flat and regresses Music below the baseline. 
Adding the continuous thinking block improves General, Emotion, and ASR, and partially recovers Music, though AIR-Bench and Music remain below the original Qwen2.5-Omni baseline. 
Representational expert distillation lifts AIR-Bench substantially, further recovers Music, and raises General.
Specialized expert distillation attains the best score on every metric, including the only Music value that surpasses the original baseline. 
Relative to SFT alone, CoAT improves every metric and is the only configuration that lifts both broad reasoning and music understanding above the original Qwen2.5-Omni baseline. 
These results show that the improvements are not driven by SFT alone: representational expert distillation accounts for the gains in general understanding and reasoning, while specialized expert distillation yields the largest improvements on task-relevant metrics.
\input{table/abl_task}
\input{table/abl_projector}

\paragraph{Projector Type.}
The choice of projector affects how the thinking representation is decoded into expert-aligned predictions. 
We compare a linear projector against the single-block Transformer projector used in our main results. Table~\ref{tab:abl_projector} shows that the Transformer projector generally outperforms the simple linear projection layer.

%% file: table/main_all.tex
\begin{table}[t]
  \caption{\textbf{Main results.} Per-benchmark performance of Qwen2-Audio, Qwen2.5-Omni-7B, and Audio Flamingo 3 with and without CoAT.}
  \label{tab:main_all}
  \centering
  \begin{adjustbox}{max width=\textwidth}
  \setlength{\tabcolsep}{3pt}
  \begin{tabular}{l |c| cc| cc| cc}
    \toprule
    Benchmark            & Eval                       & Qwen2-Audio & + \textbf{CoAT}  & Qwen2.5-Omni       & + \textbf{CoAT}  & Audio Flamingo 3    & + \textbf{CoAT} \\
    \midrule
    \midrule
    \multicolumn{8}{l}{\textit{\textbf{Audio Understanding \& Reasoning}}} \\
    \midrule
    \midrule
    \multicolumn{8}{l}{\textit{General}} \\
    \midrule
    MMAU                 & Acc\,$\uparrow$            & 52.50 & \textbf{66.90}   & 65.60 & \textbf{69.40}   & 69.40 & \textbf{70.00} \\
    MMAR                 & Acc\,$\uparrow$            & 47.10 & \textbf{52.60}   & 56.70 & \textbf{62.20}   & 55.70 & \textbf{59.60} \\
    MMSU                 & Acc\,$\uparrow$            & 53.31 & \textbf{58.27}   & 61.32 & \textbf{64.99}   & \textbf{60.01} & 58.36 \\
    ClothoAQA            & Acc\,$\uparrow$            & 75.52 & \textbf{79.96}   & 84.67 & \textbf{87.17}   & 80.10 & \textbf{85.30} \\
    Alpaca-Audio         & GPT\,$\uparrow$   & \textbf{57.40} & 56.97   & 60.60 & \textbf{64.24}   & 38.80 & \textbf{58.59} \\
    WavCaps              & GPT\,$\uparrow$     & 32.20 & \textbf{39.40}    & 29.40        & \textbf{33.60}    & \textbf{33.40} & 29.00 \\
    \midrule
    \multicolumn{8}{l}{\textit{AIR-Bench}} \\
    \midrule
    ABF Speech           & Acc\,$\uparrow$            & 50.16 & \textbf{70.98}   & 67.33 & \textbf{75.99}   & 62.99 & \textbf{71.24} \\
    ABF Sound            & Acc\,$\uparrow$            & 63.36 & \textbf{71.64}   & 76.28 & \textbf{76.49}   & 65.56 & \textbf{69.50} \\
    ABF Music            & Acc\,$\uparrow$            & 51.23 & \textbf{55.82}   & 58.20 & \textbf{59.38}   & 58.62 & \textbf{64.50} \\
    \midrule
    \multicolumn{8}{l}{\textit{Music Classification}} \\
    \midrule
    VocalSound           & Acc\,$\uparrow$ & 88.10 & \textbf{93.59}   & 91.36   & \textbf{91.70}   & \textbf{93.06} & 92.39 \\
    GTZAN                & Acc\,$\uparrow$ & 90.29 & \textbf{92.39}   & 92.89   & \textbf{92.99}   & 94.99 & \textbf{95.50} \\
    MuchoMusic           & Acc\,$\uparrow$ & 69.25 & \textbf{74.05}   & 72.87   & \textbf{73.97}   & 81.63 & \textbf{81.80} \\
    \midrule
    \multicolumn{8}{l}{\textit{Speech Emotion Recognition}} \\
    \midrule
    MELD                 & Acc\,$\uparrow$ / F1\,$\uparrow$  & 25.5 / 21.8 & \textbf{58.0} / \textbf{56.1}  & 49.4 / 51.1 & \textbf{60.8} / \textbf{59.2} & 40.8 / 45.9 & \textbf{59.8} / \textbf{57.2} \\
    IEMOCAP              & Acc\,$\uparrow$            & 54.00 & \textbf{72.70}   & 55.22 & \textbf{79.01}   & 63.58 & \textbf{70.39} \\
    \midrule
    \midrule
    \multicolumn{8}{l}{\textit{\textbf{Automatic Speech Recognition}}} \\
    \midrule
    \midrule
    LibriSpeech-clean    & \multirow{7}{*}{WER\,$\downarrow$}          & 4.14 & \textbf{2.30}    & 1.80   & \textbf{1.77}    & \textbf{1.57}   & 1.99 \\
    LibriSpeech-other    &          & 7.02 & \textbf{4.74}    & \textbf{3.40}   & 4.00    & \textbf{3.13} & 4.23 \\
    Common Voice 15      &          & 21.48 & \textbf{9.25}    & \textbf{7.60}   & 7.68    & \textbf{7.40}  & \textbf{7.40} \\
    GigaSpeech           &           & 13.53 & \textbf{10.52}    & 10.42  & \textbf{9.42}    & \textbf{10.27}  & 11.90 \\
    VoxPopuli            &           & 11.90 & \textbf{7.03}    & 5.80   & \textbf{5.56}    & \textbf{5.55} & 5.70 \\
    SPGISpeech           &          & 6.93 & \textbf{2.86}    & 2.79   & \textbf{2.10}    & 1.86  & \textbf{1.84} \\
    Switchboard          &         & 19.06 & \textbf{9.25}    & 14.83  & \textbf{7.35}    & 8.01 & \textbf{7.18} \\
    \bottomrule
  \end{tabular}
  \end{adjustbox}
\end{table}

%% file: table/comp_discrete.tex
\begin{table}[t]
    \centering
    \caption{\textbf{Comparison with discrete reasoning.} Reasoning accuracy and per-sample inference cost on the full MMAU and MMAR evaluation sets.}
    \label{tab:comp_discrete}
    \footnotesize
    \resizebox{\textwidth}{!}{
    \begin{tabular}{l cc rrrr}
        \toprule
                                  & MMAU $\uparrow$ & MMAR $\uparrow$ & TTFT (s) $\downarrow$ & Dec.~time (s) $\downarrow$ & Dec.~tok  & Total (s) $\downarrow$ \\
        \midrule
        Qwen2.5-Omni              & 65.60 & 56.70 & 0.124                & 0.014                 & 6.5                  & 0.139                \\
        \;\;w/ text-CoT (prompt)  & 66.90 & 56.70 & 0.133 ($1.07\times$) & 0.147 ($10.5\times$)  & 59.5 ($9.15\times$)  & 0.280 ($2.01\times$) \\
        \;\;w/ CoAT (ours)        & 69.40 & 62.20 & 0.126 ($1.01\times$) & 0.019 ($1.34\times$)  & 8.5 ($1.31\times$)   & 0.145 ($1.04\times$) \\
        \midrule
        Audio Flamingo 3          & 69.40 & 55.70 & 0.034                & 0.004                 & 2.2                  & 0.038                \\
        \;\;w/ text-CoT           & 64.52 & 54.25 & 0.034 ($1.00\times$) & 0.205 ($55.3\times$)  & 80.4 ($36.5\times$)  & 0.239 ($6.24\times$) \\
        \;\;w/ CoAT (ours)        & 70.00 & 59.60 & 0.037 ($1.07\times$) & 0.028 ($7.57\times$)  & 11.7 ($5.32\times$)  & 0.065 ($1.70\times$) \\
        \bottomrule
    \end{tabular}}
\end{table}

%% file: table/probe_mediation.tex
\centering
\footnotesize
\setlength{\tabcolsep}{3pt}
\renewcommand{\arraystretch}{1.05}
\begin{adjustbox}{max width=\linewidth}
\begin{tabular}{l cc cc}
  \toprule
        & \multicolumn{2}{c}{IEMOCAP} & \multicolumn{2}{c}{MuchoMusic} \\
  \cmidrule(lr){2-3} \cmidrule(lr){4-5}
  Model & Think & Pre-gen & Think & Pre-gen \\
  \midrule
  Qwen2.5-Omni                 & --                          & .622/.32                    & --                          & .510/.06 \\
  CoAT (recon+spidr)           & .630/.36                    & .707/.57                    & .301/.20                  & .567/.18                       \\
  CoAT (full)                  & \textbf{.705}/\textbf{.49}  & \textbf{.727}/\textbf{.56}  & \textbf{.324}/\textbf{.21}& \textbf{.607}/\textbf{.22}    \\
  \bottomrule
\end{tabular}
\end{adjustbox}

%% file: table/abl_task.tex
\begin{table}[t]
  \caption{\textbf{Task ablation.} Starting from the baseline model, we cumulatively add one component at a time and retrain under the same schedule: supervised fine-tuning (SFT), continuous thinking block, representational expert distillation, and specialized expert distillation. We report the average performance of benchmarks with the same metric.}
  \label{tab:abl_task}
  \centering
  \begin{adjustbox}{max width=\textwidth}
  \begin{tabular}{l ccccc}
    \toprule
    Variant
      & General (Acc\,$\uparrow$)
      & AIR-Bench (Acc\,$\uparrow$)
      & Music (Acc\,$\uparrow$)
      & Emotion (Acc\,$\uparrow$) 
      & ASR (WER\,$\downarrow$) \\
    \midrule
    Qwen2.5-Omni                & 67.07 & 67.27 & 85.71 & 52.29 & 6.66\\
    \;+ SFT & 67.48 & 67.33 & 84.65 & 67.31 & 5.58\\
    \;+ Continuous Thinking Block & 68.52 & 67.01 & 84.86 & 67.14 & 5.42\\
    \;+ Representational Expert Distillation & 69.89 & 68.75 & 84.39 & 67.32 & 5.54\\
    \;+ Specialized Expert Distillation & \textbf{70.94} & \textbf{70.62} & \textbf{86.22} & \textbf{69.91} & \textbf{5.41}  \\
    \bottomrule
  \end{tabular}
  \end{adjustbox}
\end{table}

%% file: table/abl_projector.tex
\begin{table}[t]
  \caption{\textbf{Projector type.} We compare a linear projection against the single-block Transformer projector used in the main results. We report the average performance of benchmarks with the same metric.}
  \label{tab:abl_projector}
  \centering
  \begin{adjustbox}{max width=\textwidth}
  \begin{tabular}{l ccccc}
    \toprule
    Projector
      & General (Acc\,$\uparrow$)
      & AIR-Bench (Acc\,$\uparrow$)
      & Music (Acc\,$\uparrow$)
      & Emotion (Acc\,$\uparrow$) 
      & ASR (WER\,$\downarrow$) \\
    \midrule
    Linear                          &  70.08 & \textbf{70.75} & 83.86 & 69.37 &  5.50\\
    Transformer (ours)              & \textbf{70.94} & 70.62 & \textbf{86.22} & \textbf{69.91} & \textbf{5.41}  \\
    \bottomrule
  \end{tabular}
  \end{adjustbox}
\end{table}

%% file: tex/5_con.tex
\section{Conclusion}
\label{sec:conclusion}

We introduce Continuous Audio Thinking (CoAT), a framework that equips audio language models with a continuous latent workspace for organizing acoustic information before response generation. 
Audio experts supervise this workspace through distillation under a representational-then-specialized schedule: latent states are first anchored to the audio space and then aligned with semantic, paralinguistic, and musical structure.
CoAT consistently improves audio understanding and reasoning across three LALMs, while running at lower per-sample latency than text chain-of-thought.
Further analyses show that the auxiliary signal propagates from the thinking positions to the textual outputs: linear probes on those positions became more accurate over training, and within-model probe confidence predicted downstream task performance. 
CoAT shows that continuous latent thinking can support reasoning in modalities that are difficult to verbalize.

\section*{Limitations and Future Work}
\label{app:limitations}
Our study has two main limitations, which we leave to future work. 
First, the thinking block in CoAT is deterministic, occupying a fixed span at a pre-defined position between the audio input and the assistant response. 
The model does not learn when or how long to think, nor does it interleave thinking with response generation. 
CoAT thus realizes a latent workspace but not multi-step latent reasoning, and extending it to dynamic or interleaved thinking blocks is an important next step. 
Second, our empirical validation is confined to the audio domain. 
Although the proposed mechanism is modality-agnostic in principle, whether the same recipe transfers to vision- and video-language models remains an open question.

\section*{Acknowledgements}
This work was partly supported by the InnoCORE program of the Ministry of Science and ICT (AI Meta-Scientist, N10260110) and Artificial intelligence industrial convergence cluster development project funded by the Ministry of Science and ICT(MSIT, Korea)\&Gwangju Metropolitan City. All experiments were conducted on Cosmo, PYLER's internal GPU cluster.

%% file: tex/appendix.tex
\appendix
\setcounter{table}{0}
\renewcommand{\thetable}{\Alph{table}}

\setcounter{figure}{0}
\renewcommand{\thefigure}{\Alph{figure}}

\section{Experimental Details}
\label{app:configs}

\paragraph{Training configuration.}
All three backbones produce audio token embeddings at $r_s = 25$\,Hz, with LM hidden size $d = 4096$ for Qwen2-Audio and $d = 3584$ for Qwen2.5-Omni-7B and Audio Flamingo 3. Throughout the main CoAT training runs, the audio tower and all expert encoders are kept frozen. Gradients update only the LoRA adapters and the per-task projection heads. We use LoRA adapters of rank $16$ with $\alpha = 32$. Each projection head $P_k$ is a single-block Transformer with multi-head attention, a feed-forward layer, and a linear map to the corresponding expert embedding dimension.

CoAT is trained with a two-stage schedule. The first stage is a reconstruction-only warm-up that aligns the thinking states with the audio-token distribution. The second stage activates all five expert losses together with the language-modeling loss. The loss weight for each expert compensates for differences in output scale, and the same optimizer, learning rate, LoRA configuration, and training schedule are used for every backbone. Training runs on $4\!\times$~NVIDIA~B200 GPUs with an effective batch size of $16$, taking approximately $88$~B200 GPU-hours per backbone for the full two-stage schedule. The full evaluation suite requires approximately $15$~B200 GPU-hours per evaluated model. Table~\ref{tab:teachers} lists the expert encoders, and Table~\ref{tab:hyperparams} lists the training schedule and hyperparameters.

\input{table/teachers}

\input{table/training_hyperparameter}

\input{table/dataset}

\paragraph{Training datasets.}
Training data are drawn from publicly available sources organized into seven task groups: automatic speech recognition, audio and speech question answering, audio captioning, multiple-choice audio understanding, music understanding, spoken-instruction following, and a small text-only supervised fine-tuning split. The text-only split is included to preserve instruction-following ability and to prevent catastrophic over-refusal on harmful prompts, we observed that both abilities degraded rapidly under audio-only multi-task training. After fixing the per-task sampling ratio, we draw a training subset sized so that the full step schedule corresponds to one epoch over the subset. To prevent overfitting on smaller tasks, we additionally cap the maximum number of epochs that any single task can repeat within this subset.

Table~\ref{tab:corpus} lists the full training corpus together with the sampling ratios used to materialize a single stage of $1{,}600{,}000$ shuffled rows. Each category draws independently from its constituent sources at uniform probability, so per source budgets within a category scale with pool size. The number of optimization steps is configured so that effective batch size multiplied by total steps matches the stage size, giving each row in the materialized file approximately one expected pass per stage and avoiding within stage oversampling. The spoken instruction following category pools the evaluation aligned GSQA prompts and the text only WildJailbreak split is included to anchor refusal behavior on harmful prompts while preventing over refusal on benign adversarial ones.

\paragraph{Evaluation suite.}
We evaluate CoAT on a broad benchmark suite organized into five families that mirror the rows of Table~\ref{tab:main_all}. General audio understanding and reasoning uses MMAU~\citep{mmau}, MMAR~\citep{mmar}, MMSU~\citep{mmsu}, ClothoAQA~\citep{clothoaqa}, Alpaca-Audio~\citep{alpacaaudio}, and WavCaps~\citep{wavcaps}. AIR-Bench Foundation~\citep{airbench} contributes its three subsets covering speech, sound, and music. Music classification spans VocalSound~\citep{vocalsound}, GTZAN~\citep{gtzan}, and MuchoMusic~\citep{muchomusic}. Speech emotion recognition is measured on MELD~\citep{meld} and IEMOCAP~\citep{iemocap}. Speech transcription uses LibriSpeech~\citep{librispeech}, Common Voice 15~\citep{commonvoice}, GigaSpeech~\citep{gigaspeech}, VoxPopuli~\citep{voxpopuli}, SPGISpeech~\citep{spgispeech}, and Switchboard~\citep{switchboard}. MELD reports both accuracy and class-support-weighted F1, following the convention of the MELD paper for class-imbalanced 7-way emotion classification.

\section{Evaluation Protocol Details}
\label{app:eval-protocol}

All inference uses vLLM as the backend with greedy decoding
($T = 0$, top-$p = 0.95$). For tasks supplied by upstream
\texttt{lmms-eval}~\citep{lmmseval} we adopt the released
implementation as-is. For benchmarks that are either not natively
supported by upstream or whose upstream implementation we found to
mis-handle the audio, the prompt template, or the metric, we use our
own task definitions. These custom definitions cover IEMOCAP, MELD, SPGISpeech,
Switchboard, GTZAN, and VocalSound.

MELD's seven emotion classes are highly imbalanced, with neutral alone accounting for roughly half of the test set, so accuracy alone can be inflated by a model that always predicts the majority class. We therefore report a weighted F1 score, defined as the average of per-class F1 with each class weighted by its support (\textit{i.e.}, the number of test samples in that class).

\begin{table}[t]
\centering
\caption{\textbf{Per-benchmark evaluation protocol.} $M$ denotes
\texttt{max\_new\_tokens} as configured in \texttt{lmms-eval}. LLM
judges use the official prompt of each benchmark, reproduced verbatim.} 
\label{tab:eval-protocol}
\small
\setlength{\tabcolsep}{3pt}
\begin{tabular}{llccl}
\toprule
Family & Benchmark & Split & $M$ & Metric / Judge \\
\midrule
\multirow{6}{*}{General Audio}
 & MMAU                      & test\_mini         & 128  & Accuracy \\
 & MMAR                            & test               & 128  & Accuracy \\
 & MMSU                            & test               & 256  & Accuracy \\
 & ClothoAQA                       & test               & 8  & Exact match \\
 & Alpaca-Audio                    & test               & 1024 & GPT-4o \\
 & WavCaps                         & test               & 1024  & GPT-4o \\
\midrule
\multirow{3}{*}{AIR-Bench Foundation}
 & ABF Speech & foundation         & 256 & Accuracy \\
  & ABF Sound & foundation         & 256 & Accuracy \\
   & ABF Music & foundation         & 256 & Accuracy \\
\midrule
\multirow{3}{*}{Music classification}
 & VocalSound     & test               & 20 & Accuracy \\
 & GTZAN           & test               & 20   & Accuracy \\
 & MuchoMusic                      & test               & 256  & Accuracy \\
\midrule
\multirow{2}{*}{Speech emotion recognition}
 & MELD           & test               & 20   & Accuracy / F1 \\
 & IEMOCAP         & Session 5 (LOSO)   & 20   & Accuracy \\
\midrule
\multirow{6}{*}{Automatic Speech Recognition}
 & LibriSpeech-clean / -other      & test               & 256 & WER $\downarrow$ \\
 & Common Voice 15                 & test (en)          & 256  & WER $\downarrow$ \\
 & GigaSpeech                      & test               & 256  & WER $\downarrow$ \\
 & VoxPopuli                       & test (en)          & 2048 & WER $\downarrow$ \\
 & SPGISpeech     & test               & 512  & WER $\downarrow$ \\
 & Switchboard     & eval2000           & 512  & WER $\downarrow$ \\
\bottomrule
\end{tabular}
\end{table}

Audio clips longer than $120$ seconds are skipped at data-loading. This affects only a negligible fraction of inputs and is applied identically to every model, so it does not affect relative comparisons. GigaSpeech references additionally contain meta-tags (\texttt{<MUSIC>}, \texttt{<NOISE>}, \texttt{<SIL>}, \texttt{<OTHER>}) that are stripped before scoring. Samples whose reference becomes empty after this stripping are excluded from the WER aggregate, since an empty reference is unscorable and would otherwise count every emitted word as a pure insertion error. No analogous empty-reference filter is applied to the other ASR benchmarks, which do not use such tags.

\begin{figure}[!ht]
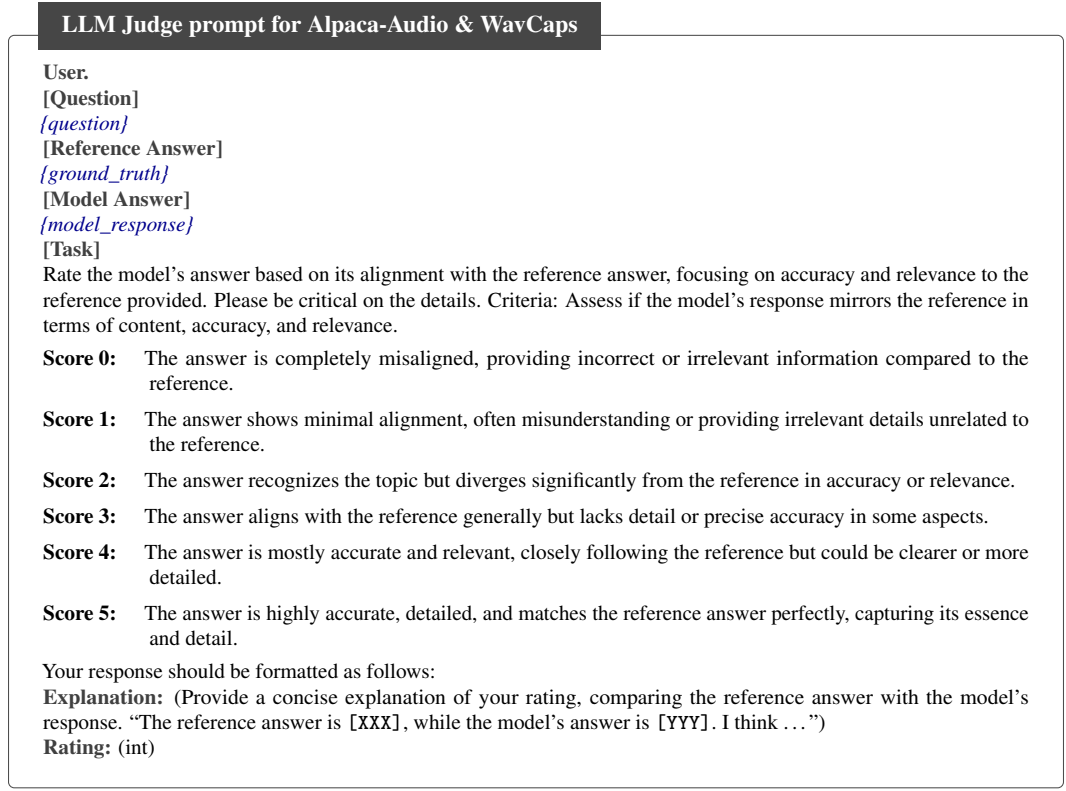

\footnotesize
{\bfseries Alpaca-Audio \& WavCaps\, (single response, 0--100)}
\par\medskip
\noindent\textit{Judge model:} \texttt{gpt-4o-2024-11-20}\;\textbar\;
\textit{Decoding:} temperature $0$, max tokens $1024$\;\textbar\;
\textit{System message:} none\;\textbar\;
\textit{Variables:} \promptvar{\{question\}},
\promptvar{\{ground\_truth\}}, \promptvar{\{model\_response\}}.

\vspace{0.5em}
\begin{promptbox}{LLM Judge prompt for Alpaca-Audio \& WavCaps}
\promptsec{User.}\\
\promptsec{[Question]}\\
\promptvar{\{question\}}

\promptsec{[Reference Answer]}\\
\promptvar{\{ground\_truth\}}

\promptsec{[Model Answer]}\\
\promptvar{\{model\_response\}}

\promptsec{[Task]}\\
Rate the model's answer based on its alignment with the reference
answer, focusing on accuracy and relevance to the reference provided.
Please be critical on the details.
Criteria: Assess if the model's response mirrors the reference in
terms of content, accuracy, and relevance.

\begin{description}[
        labelwidth=4.2em, leftmargin=5em, labelsep=0.6em,
        font=\normalfont\bfseries, itemsep=2pt, topsep=2pt,
        align=left]
  \item[Score 0:] The answer is completely misaligned, providing
        incorrect or irrelevant information compared to the reference.
  \item[Score 1:] The answer shows minimal alignment, often
        misunderstanding or providing irrelevant details unrelated to
        the reference.
  \item[Score 2:] The answer recognizes the topic but diverges
        significantly from the reference in accuracy or relevance.
  \item[Score 3:] The answer aligns with the reference generally but
        lacks detail or precise accuracy in some aspects.
  \item[Score 4:] The answer is mostly accurate and relevant, closely
        following the reference but could be clearer or more detailed.
  \item[Score 5:] The answer is highly accurate, detailed, and matches
        the reference answer perfectly, capturing its essence and
        detail.
\end{description}

Your response should be formatted as follows:\\
\promptsec{Explanation:} (Provide a concise explanation of your
rating, comparing the reference answer with the model's response.
``The reference answer is \texttt{[XXX]}, while the model's answer is
\texttt{[YYY]}. I think \dots'')\\
\promptsec{Rating:} (int)
\end{promptbox}
\caption{LLM judge configuration and prompt for Alpaca-Audio \& WavCaps.}
\label{fig:judge-prompt}
\end{figure}

Figure~\ref{fig:judge-prompt} shows the LLM judge configuration and prompt used for WavCaps and Alpaca-Audio, with model and decoding hyperparameters listed above the prompt box. Inside the box, \promptsec{System.} and \promptsec{User.} mark the chat roles, \promptsec{[Section]} markers are part of the literal prompt, and \promptvar{italic blue} text denotes per-sample variables. We normalized the 5-point scale scores to a 100-point scale.

\section{Inference Cost Analysis}
\label{app:speed-duration}
\input{table/comp_discrete_duration}
\paragraph{Setup.} Two complementary measurements support Table~\ref{tab:comp_discrete}.
For the main table, we run all variants on the full MMAU and MMAR evaluation sets ($2{,}000$ samples total) and report per-sample mean latency.
For the duration-stratified analysis, we draw a fixed $300$ samples per audio-duration range from the broader evaluation suite, which is otherwise heavily skewed toward short audio with over 90\% of samples below 15\,s, giving $1{,}500$ samples spanning 0 to 120\,s, and repeat the sweep with three independent random seeds.
Inference is run with batch size $1$ on $4\times$NVIDIA~B200 (tensor parallel size $4$) under vLLM~0.19, with prefix caching disabled so that every request pays its full prefill cost.
Note that the performance of the AF3-think model is evaluated with GPT-eval due to the answer parsing problem.

\paragraph{Per-duration results.} Table~\ref{tab:comp_discrete_duration} reports time to first token, decode time, total wall-clock, decoded tokens, and CoAT's forced-prefix length for each audio-duration range as mean $\pm$ std across 3 random seeds. Two patterns stand out.
First, CoAT's forced-prefix length grows monotonically with audio duration and spans more than an order of magnitude across the five ranges. Despite this growth, CoAT's time to first token stays within a few milliseconds of the same-backbone baseline in every range, indicating that the additional prefill compute is largely absorbed by the audio encoder's existing cost.
Second, text-CoT's decoded-token count stays roughly flat across all five ranges, reflecting that the textual chain-of-thought is determined by task format rather than audio duration. Its decode wall-clock is correspondingly stable across ranges and contributes the dominant cost regardless of input length.

\section{Inference-Time Think Token vs.\ CoAT}
\label{app:af3-think}

\input{table/af3_think_all}

A natural alternative to CoAT is to ask whether the gains we attribute to our reconstruction teachers can in fact be obtained for free, simply by allocating extra inference-time compute to a think segment.
On the Audio Flamingo 3 backbone we isolate this question by comparing two settings: an inference-time think policy that prepends a think segment to the frozen pretrained model, and our CoAT fine-tune that supervises the same segment with multi-teacher reconstruction targets.
Table~\ref{tab:af3_think_all} reports per-benchmark scores together with the gap $\Delta$ between CoAT and another model.

The Audio Flamingo 3 denotes the vanilla AF3 model, whereas “think” or “with CoT” refers the AF3 with additional thinking. The two policies behave very differently.
Inference-time thinking by itself is largely neutral or harmful, regressing on a clear majority of benchmarks across audio understanding, music, and emotion families, with the largest drops concentrated on captioning and open-ended QA.
CoAT, in contrast, delivers consistent gains on the metrics that exercise audio understanding---most prominently in voice-assistant style QA, AIR-Bench Foundation, and speech emotion recognition---at the cost of a modest WER regression on most ASR test sets, consistent with mild distribution shift from fine-tuning on a multi-task corpus.
The takeaway is that the benefit of CoAT comes from what the think segment is supervised to reconstruct, not from the segment's mere presence at inference.

\section{Societal Impact}
\label{app:societal_impact}

CoAT improves audio understanding in audio language models, with potential positive applications such as accessibility (captioning, transcription, assistive listening) and content moderation.
As with audio language models in general, the same improvements may have dual-use implications that warrant consideration at deployment time. These considerations apply at the level of the underlying backbone rather than being introduced by CoAT itself.

%% file: table/teachers.tex
\begin{table}[t]
  \caption{Audio expert encoders used in CoAT. $e_k$ is the expert embedding dimension and $r_k$ is the frame rate at which the expert emits features.}
  \label{tab:teachers}
  \centering
  \begin{tabular}{llccc}
    \toprule
    Expert & Aspect & $e_k$ & $r_k$ & Loss \\
    \midrule
    Sim-Whisper~\citep{simwhisper} & self-reconstruction & 1280 & 25\,Hz & MSE \\
    SPIDR~\citep{poli2025spidr} & speech & 768 & 50\,Hz & MSE \\
    PANNs~\citep{kong2020panns} & sound events & 2048 & 25\,Hz & BCE \\
    emotion2vec~\citep{ma2024emotion2vec} & paralinguistic & 1024 & 50\,Hz & MSE \\
    basic\_pitch~\citep{basicpitch} & pitch & 2112 & 86\,Hz & MSE + Focal BCE \\
    \bottomrule
  \end{tabular}
\end{table}

%% file: table/training_hyperparameter.tex
\begin{table}[t]
\centering
\caption{\textbf{Training schedule.} CoAT is trained in two stages
that share the same optimizer, learning rate, and batch configuration.
Only the active expert set and the per-expert loss weights change.
Stage~1 is a reconstruction-only warm-up, and stage~2 activates all
five experts. The basic\_pitch expert carries two coupled losses,
an MSE on its dense intermediate feature and an auxiliary focal-BCE
on the 264-bin pitch posteriorgram.}
\label{tab:hyperparams}
\small
\begin{tabular}{lcc}
\toprule
 & Stage 1 (warm-up) & Stage 2 (multi-task) \\
\midrule
\multicolumn{3}{l}{\textit{Schedule and optimization}} \\
Steps                                  & 20{,}000             & 80{,}000 \\
Learning rate                          & $5\times10^{-5}$     & $5\times10^{-5}$ \\
LR schedule                            & linear, warmup $0.03$ & linear, warmup $0.03$ \\
Optimizer                              & AdamW                & AdamW \\
Precision                              & bf16                 & bf16 \\
Per-device batch / grad-accum          & $4 / 1$              & $4 / 1$ \\
Effective batch (4 GPUs)               & $16$  & $16$ \\
Max sequence length                    & $2048$               & $2048$ \\
LoRA $r$ / $\alpha$ / dropout          & $16 / 32 / 0.05$     & $16 / 32 / 0.05$ \\
Expert dropout $p_{\text{drop}}$      & --                   & $0.5$ \\
\midrule
\multicolumn{3}{l}{\textit{Active experts and loss weights $w_k$}} \\
$w_{\text{recon}}$ (Sim-Whisper, MSE)              & $5.0$ & $2.0$ \\
$w_{\text{spidr}}$ (SPIDR, MSE)                    & --    & $0.2$ \\
$w_{\text{uss}}$ (PANNs USS, BCE)                  & --    & $100$ \\
$w_{\text{emotion}}$ (emotion2vec, MSE)            & --    & $40$ \\
$w_{\text{pitch}}$ (basic\_pitch feature, MSE)     & --    & $5.0$ \\
$w_{\text{pitch}}^{\text{aux}}$ (basic\_pitch posteriorgram, focal-BCE) & -- & $8$ \\
\bottomrule
\end{tabular}
\end{table}

%% file: table/dataset.tex
\begin{table}[t]
\centering
\caption{Training corpus and single-stage sampling budget (1.6M target samples). Pool is the unique row count per source. Subtotal aggregates the pool within each category. Share is the per-category mixing ratio; Samples is the resulting sample budget.}
\label{tab:corpus}
\small
\setlength{\tabcolsep}{4pt}
\renewcommand{\arraystretch}{1.05}
\begin{adjustbox}{max width=\textwidth}
\begin{tabular}{l l r r r r}
\toprule
\textbf{Category} & \textbf{Source dataset} & \textbf{Pool} & \textbf{Subtotal} & \textbf{Share} & \textbf{Samples} \\
\midrule
\multirow{6}{*}{Automatic Speech Recognition}
 & LibriSpeech         &   281{,}238 & \multirow{6}{*}{2{,}699{,}934} & \multirow{6}{*}{21.25\%} & \multirow{6}{*}{340{,}000} \\
 & GigaSpeech          &   910{,}122 & & & \\
 & CommonVoice 15 (en) & 1{,}070{,}055 & & & \\
 & VoxPopuli (en)      &   182{,}466 & & & \\
 & Switchboard         &   178{,}980 & & & \\
 & SPGISpeech          &    77{,}073 & & & \\
\midrule
\multirow{4}{*}{Audio \& speech QA}
 & OpenASQA--AudioCaps & 534{,}473 & \multirow{4}{*}{758{,}560} & \multirow{4}{*}{21.25\%} & \multirow{4}{*}{340{,}000} \\
 & OpenASQA--IEMOCAP   & 108{,}497 & & & \\
 & LibriSQA (open)     & 103{,}928 & & & \\
 & Clotho-AQA          &  11{,}662 & & & \\
\midrule
\multirow{2}{*}{Audio captioning}
 & AudioCaps & 45{,}178 & \multirow{2}{*}{51{,}107} & \multirow{2}{*}{14.00\%} & \multirow{2}{*}{224{,}000} \\
 & Clotho-v2 &  5{,}929 & & & \\
\midrule
\multirow{3}{*}{\shortstack[l]{Multiple-choice\\audio understanding}}
 & MELD (emotion MCQ)    &   9{,}988 & \multirow{3}{*}{121{,}677} & \multirow{3}{*}{23.50\%} & \multirow{3}{*}{376{,}000} \\
 & IEMOCAP (emotion MCQ) &   7{,}676 & & & \\
 & LibriSQA (MCQ)        & 104{,}013 & & & \\
\midrule
\multirow{2}{*}{Music understanding}
 & MusicBench (caption) &  52{,}768 & \multirow{2}{*}{105{,}536} & \multirow{2}{*}{10.00\%} & \multirow{2}{*}{160{,}000} \\
 & MusicBench (MCQ)     &  52{,}768 & & & \\
\midrule
\multirow{1}{*}{Spoken-instruction following}
 & GSQA  &  51{,}349 & 51{,}349 & 5.00\% & 80{,}000 \\
\midrule
Text-only SFT
 & WildJailbreak & 261{,}538 & 261{,}538 & 5.00\% & 80{,}000 \\
\midrule
\textbf{Total} & & & \textbf{4{,}049{,}701} & \textbf{100.00\%} & \textbf{1{,}600{,}000} \\
\bottomrule
\end{tabular}
\end{adjustbox}
\end{table}

%% file: table/comp_discrete_duration.tex
\begin{table*}[t]
\centering
\caption{\textbf{Inference cost by audio duration.} Per-duration mean $\pm$ std across 3 random seeds, with 300 samples per range per seed. TTFT, decode, and total are reported in seconds. Forced prefix is the length in tokens of CoAT's prepended thinking block. ``--'' marks variants that do not prepend a thinking block. Hardware and serving configuration follow Appendix~\ref{app:speed-duration}.}
\label{tab:comp_discrete_duration}
\setlength{\tabcolsep}{4pt}
\small
\begin{tabular}{ll rrr r r}
\toprule
Duration & Model & TTFT $\downarrow$ & Decode $\downarrow$ & Total $\downarrow$ & Dec.~tok $\downarrow$ & Forced prefix \\
\midrule
\multirow{6}{*}{0--5\,s}
 & Qwen2.5-Omni        & 0.136 $\pm$ 0.004 & 0.036 $\pm$ 0.004 & 0.172 $\pm$ 0.001 & 14.9 $\pm$ 1.7  & --             \\
 & \;\;w/ text-CoT     & 0.142 $\pm$ 0.009 & 0.075 $\pm$ 0.007 & 0.217 $\pm$ 0.003 & 30.5 $\pm$ 2.8  & --             \\
 & \;\;w/ CoAT         & 0.147 $\pm$ 0.010 & 0.023 $\pm$ 0.002 & 0.170 $\pm$ 0.012 & 9.8 $\pm$ 0.8   & 71 $\pm$ 1     \\
 & Audio Flamingo 3    & 0.029 $\pm$ 0.000 & 0.015 $\pm$ 0.001 & 0.044 $\pm$ 0.000 & 6.5 $\pm$ 0.3   & --             \\
 & \;\;w/ text-CoT     & 0.031 $\pm$ 0.004 & 0.151 $\pm$ 0.008 & 0.183 $\pm$ 0.010 & 59.8 $\pm$ 3.3  & --             \\
 & \;\;w/ CoAT         & 0.029 $\pm$ 0.000 & 0.021 $\pm$ 0.001 & 0.051 $\pm$ 0.001 & 9.3 $\pm$ 0.4   & 71 $\pm$ 1     \\
\midrule
\multirow{6}{*}{5--15\,s}
 & Qwen2.5-Omni        & 0.131 $\pm$ 0.005 & 0.070 $\pm$ 0.006 & 0.201 $\pm$ 0.009 & 28.9 $\pm$ 2.2  & --             \\
 & \;\;w/ text-CoT     & 0.139 $\pm$ 0.006 & 0.091 $\pm$ 0.002 & 0.230 $\pm$ 0.007 & 37.1 $\pm$ 0.6  & --             \\
 & \;\;w/ CoAT         & 0.134 $\pm$ 0.004 & 0.063 $\pm$ 0.001 & 0.198 $\pm$ 0.004 & 26.1 $\pm$ 0.4  & 220 $\pm$ 4    \\
 & Audio Flamingo 3    & 0.029 $\pm$ 0.000 & 0.054 $\pm$ 0.003 & 0.083 $\pm$ 0.003 & 22.0 $\pm$ 1.1  & --             \\
 & \;\;w/ text-CoT     & 0.029 $\pm$ 0.001 & 0.176 $\pm$ 0.004 & 0.206 $\pm$ 0.004 & 70.0 $\pm$ 1.3  & --             \\
 & \;\;w/ CoAT         & 0.033 $\pm$ 0.000 & 0.062 $\pm$ 0.003 & 0.095 $\pm$ 0.003 & 25.3 $\pm$ 1.2  & 220 $\pm$ 4    \\
\midrule
\multirow{6}{*}{15--30\,s}
 & Qwen2.5-Omni        & 0.134 $\pm$ 0.004 & 0.121 $\pm$ 0.008 & 0.255 $\pm$ 0.007 & 49.2 $\pm$ 3.3  & --             \\
 & \;\;w/ text-CoT     & 0.142 $\pm$ 0.005 & 0.164 $\pm$ 0.005 & 0.306 $\pm$ 0.006 & 66.2 $\pm$ 1.9  & --             \\
 & \;\;w/ CoAT         & 0.134 $\pm$ 0.008 & 0.046 $\pm$ 0.007 & 0.180 $\pm$ 0.002 & 18.9 $\pm$ 2.8  & 557 $\pm$ 2    \\
 & Audio Flamingo 3    & 0.035 $\pm$ 0.001 & 0.023 $\pm$ 0.007 & 0.058 $\pm$ 0.006 & 9.6 $\pm$ 2.6   & --             \\
 & \;\;w/ text-CoT     & 0.036 $\pm$ 0.001 & 0.183 $\pm$ 0.002 & 0.218 $\pm$ 0.003 & 72.1 $\pm$ 0.8  & --             \\
 & \;\;w/ CoAT         & 0.038 $\pm$ 0.000 & 0.039 $\pm$ 0.012 & 0.077 $\pm$ 0.012 & 15.9 $\pm$ 4.6  & 557 $\pm$ 2    \\
\midrule
\multirow{6}{*}{30--60\,s}
 & Qwen2.5-Omni        & 0.142 $\pm$ 0.007 & 0.043 $\pm$ 0.005 & 0.185 $\pm$ 0.012 & 17.8 $\pm$ 2.0  & --             \\
 & \;\;w/ text-CoT     & 0.150 $\pm$ 0.009 & 0.175 $\pm$ 0.006 & 0.325 $\pm$ 0.013 & 70.5 $\pm$ 2.2  & --             \\
 & \;\;w/ CoAT         & 0.143 $\pm$ 0.009 & 0.036 $\pm$ 0.004 & 0.179 $\pm$ 0.013 & 14.9 $\pm$ 1.6  & 839 $\pm$ 3    \\
 & Audio Flamingo 3    & 0.048 $\pm$ 0.001 & 0.014 $\pm$ 0.003 & 0.062 $\pm$ 0.004 & 6.2 $\pm$ 1.3   & --             \\
 & \;\;w/ text-CoT     & 0.048 $\pm$ 0.001 & 0.157 $\pm$ 0.006 & 0.205 $\pm$ 0.007 & 61.9 $\pm$ 2.3  & --             \\
 & \;\;w/ CoAT         & 0.050 $\pm$ 0.000 & 0.044 $\pm$ 0.004 & 0.093 $\pm$ 0.004 & 17.8 $\pm$ 1.4  & 839 $\pm$ 3    \\
\midrule
\multirow{6}{*}{60--120\,s}
 & Qwen2.5-Omni        & 0.183 $\pm$ 0.007 & 0.017 $\pm$ 0.001 & 0.201 $\pm$ 0.007 & 7.7 $\pm$ 0.3   & --             \\
 & \;\;w/ text-CoT     & 0.160 $\pm$ 0.010 & 0.128 $\pm$ 0.003 & 0.288 $\pm$ 0.011 & 52.3 $\pm$ 1.0  & --             \\
 & \;\;w/ CoAT         & 0.193 $\pm$ 0.009 & 0.012 $\pm$ 0.000 & 0.205 $\pm$ 0.009 & 5.6 $\pm$ 0.0   & 2034 $\pm$ 62  \\
 & Audio Flamingo 3    & 0.075 $\pm$ 0.001 & 0.004 $\pm$ 0.001 & 0.079 $\pm$ 0.001 & 2.2 $\pm$ 0.2   & --             \\
 & \;\;w/ text-CoT     & 0.074 $\pm$ 0.002 & 0.142 $\pm$ 0.002 & 0.217 $\pm$ 0.002 & 55.8 $\pm$ 0.6  & --             \\
 & \;\;w/ CoAT         & 0.083 $\pm$ 0.002 & 0.021 $\pm$ 0.005 & 0.104 $\pm$ 0.003 & 8.9 $\pm$ 1.8   & 2034 $\pm$ 62  \\
\bottomrule
\end{tabular}
\end{table*}

%% file: table/af3_think_all.tex
\begin{table}[t]
  \caption{\textbf{AF3 think-token comparison.} Per-benchmark performance of Audio Flamingo 3 with the inference-time + think token policy and our + CoAT fine-tune. Since AF3 think model does not support adaptive thinking mode, we cannot evaluate ASR with WER and therefore omit this value.}
  \label{tab:af3_think_all}
  \centering
  \small
  \begin{tabular}{l |c| c c c |c}
    \toprule
    Benchmark            & Eval                       & Audio Flamingo 3       & + think  & + CoAT & $\Delta$\\
    \midrule
    \midrule
    \multicolumn{6}{l}{\textit{\textbf{Audio Understanding \& Reasoning}}} \\
    \midrule
    \midrule
    \multicolumn{6}{l}{\textit{General}} \\
    \midrule
    MMAU                 & Acc\,$\uparrow$            & 69.40          & 64.52          & \textbf{70.00} & \textcolor{ForestGreen}{$+0.60$} / \textcolor{ForestGreen}{$+5.48$} \\
    MMAR                 & Acc\,$\uparrow$            & 55.70          & 54.25          & \textbf{59.60} & \textcolor{ForestGreen}{$+3.90$} / \textcolor{ForestGreen}{$+5.35$} \\
    MMSU                 & Acc\,$\uparrow$            & \textbf{60.01} & 57.03          & 58.36          & \textcolor{BrickRed}{$-1.65$} / \textcolor{ForestGreen}{$+1.33$} \\
    ClothoAQA            & Acc\,$\uparrow$            & 80.10          & 68.79          & \textbf{85.30} & \textcolor{ForestGreen}{$+5.20$} / \textcolor{ForestGreen}{$+16.51$} \\
    Alpaca-Audio         & GPT\,$\uparrow$   & 38.80          & 30.10          & \textbf{58.59} & \textcolor{ForestGreen}{$+19.79$} / \textcolor{ForestGreen}{$+28.49$} \\
    WavCaps              & GPT\,$\uparrow$     & \textbf{33.40}  & 41.00           & 29.00           & \textcolor{BrickRed}{$-4.40$} / \textcolor{BrickRed}{$-12.00$} \\
    \midrule
    \multicolumn{6}{l}{\textit{AIR-Bench Foundation}} \\
    \midrule
    ABF Speech           & Acc\,$\uparrow$            & 62.99          & 58.04          & \textbf{71.24} & \textcolor{ForestGreen}{$+8.25$} / \textcolor{ForestGreen}{$+13.20$} \\
    ABF Sound            & Acc\,$\uparrow$            & 65.56          & 63.39          & \textbf{69.50} & \textcolor{ForestGreen}{$+3.94$} / \textcolor{ForestGreen}{$+6.11$} \\
    ABF Music            & Acc\,$\uparrow$            & 58.62          & 54.40          & \textbf{64.50} & \textcolor{ForestGreen}{$+5.88$} / \textcolor{ForestGreen}{$+10.10$} \\
    \midrule
    \multicolumn{6}{l}{\textit{Music Classification}} \\
    \midrule
    VocalSound           & Acc\,$\uparrow$            & \textbf{93.06} & 82.22          & 92.39          & \textcolor{BrickRed}{$-0.67$} / \textcolor{ForestGreen}{$+10.17$} \\
    GTZAN                & Acc\,$\uparrow$            & 94.99          & 92.89          & \textbf{95.50} & \textcolor{ForestGreen}{$+0.51$} / \textcolor{ForestGreen}{$+2.61$} \\
    MuchoMusic           & Acc\,$\uparrow$            & 81.63          & 75.76          & \textbf{81.80} & \textcolor{ForestGreen}{$+0.17$} / \textcolor{ForestGreen}{$+6.04$} \\
    \midrule
    \multicolumn{6}{l}{\textit{Speech Emotion Recognition}} \\
    \midrule
    MELD                 & ACC \,$\uparrow$             & 40.75          & 40.46          & \textbf{59.83} & \textcolor{ForestGreen}{$+19.08$} / \textcolor{ForestGreen}{$+19.47$} \\
    
    MELD                 & F1\,$\uparrow$             & 45.93          &  44.96          & \textbf{57.16} & \textcolor{ForestGreen}{$+11.23$} / \textcolor{ForestGreen}{$+12.20$} \\
    IEMOCAP              & Acc\,$\uparrow$            & 63.58          & 56.98          & \textbf{70.39} & \textcolor{ForestGreen}{$+6.81$} / \textcolor{ForestGreen}{$+13.41$} \\
    \midrule
    \midrule
    \multicolumn{6}{l}{\textit{\textbf{Automatic Speech Recognition}}} \\
    \midrule
    \midrule
    LibriSpeech-clean    & \multirow{7}{*}{WER\,$\downarrow$} & \textbf{1.57 }         & - & 1.99          & \textcolor{BrickRed}{$+0.42$} / - \\
    LibriSpeech-other    &                                    & \textbf{3.13} & -          & 4.23          & \textcolor{BrickRed}{$+1.10$} / -\\
    Common Voice 15      &                                    & \textbf{7.40} & -         & \textbf{7.40} & $+0.00$ / - \\
    GigaSpeech           &                                    & \textbf{10.27}          & - & 11.90         & \textcolor{BrickRed}{$+1.63$} / - \\
    VoxPopuli            &                                    & \textbf{5.55} & -           & 5.70          & \textcolor{BrickRed}{$+0.15$} / - \\
    SPGISpeech           &                                    & 1.86           & -          & \textbf{1.84} & \textcolor{ForestGreen}{$-0.02$} / - \\
    Switchboard          &                                    & 8.01           & -           & \textbf{7.18} & \textcolor{ForestGreen}{$-0.83$} / - \\
    \bottomrule
  \end{tabular}
\end{table}